%% file: main.tex
\documentclass[10pt, a4paper]{article}

\usepackage{enumitem}
\usepackage{svg}
\usepackage{booktabs}
\usepackage[final]{lrec2026} % this is the new style
\title{What Makes a Good Doctor Response? A Study on Text-Based Telemedicine}

\name{Adrian Cosma$^\dagger$, Cosmin Dumitrache$^\ddagger$, Emilian Radoi$^\ddagger$} 

\address{
    $^\dagger$IDSIA, Dalle Molle Institute for Artificial Intelligence, Lugano, CH\\
    $^\ddagger$University Politehnica of Bucharest, Bucharest, RO \\
     \textbf{Correspondence}: emilian.radoi@upb.ro\\}

\abstract{
Text-based telemedicine has become an increasingly used mode of care, requiring clinicians to deliver medical advice clearly and effectively in writing. As platforms increasingly rely on patient ratings and feedback, clinicians face growing pressure to maintain satisfaction scores, even though these evaluations often reflect communication quality more than clinical accuracy. We analyse patient satisfaction signals in Romanian text-based telemedicine. Using a sample of anonymised text-based telemedicine consultations, we model feedback as a binary outcome, treating thumbs-up responses as positive and grouping negative or absent feedback into the other class. We extract from doctor responses interpretable, predominantly language-agnostic features (e.g., length, structural characteristics, readability proxies), along with Romanian LIWC psycholinguistic features and politeness/hedging markers where available. We train a classifier with a time-based split and perform SHAP-based analyses, which indicate that metadata dominates prediction, functioning as a strong prior, while characteristics of the response text provide a smaller but actionable signal. In subgroup correlation analyses, politeness and hedging are consistently associated with positive patient feedback, whereas lexical diversity shows a negative association.
\\ \newline }

\begin{document}

\maketitleabstract

\section{Introduction}
\input{sections/1.intro}

\section{Related Work}
\input{sections/2.related}

\section{Method}
\input{sections/3.method}

\section{Results}
\input{sections/4.results}

\section{Conclusions}
\input{sections/5.conclusions}

\section{Bibliographical References}
\bibliographystyle{lrec2026-natbib}
\bibliography{refs}

\end{document}

%% file: sections/1.intro.tex
Text-based telemedicine has gradually evolved from a niche offering to a more widely used way of receiving care. As more consultations move online, clinicians are increasingly practising medicine in writing. They must give clinically sound advice while explaining it clearly, using language that patients can easily understand and act upon. 
This change also introduces expectations that have emerged from other online services, especially regarding transparency and feedback. Patients are accustomed to ratings and reviews in other digital services, and many now expect similar indicators when deciding whether to trust a clinician on a telemedicine platform.

As a result, many telemedicine platforms now collect structured feedback after each consultation (e.g., thumbs up / thumbs down or star ratings, optionally accompanied by free-text comments). In contrast to in-person care, where feedback is relatively rare and often informal, online platforms can generate large volumes of explicit evaluations for individual clinicians. 

While review systems have become ubiquitous across online platforms, applying them to telemedicine consultations presents unique challenges. Doctors find themselves evaluated not by medical peers, but by patients who generally lack the clinical expertise to assess medical accuracy. This creates a fundamental tension: patient satisfaction and review scores are often driven more by how medical advice is communicated and framed than by its clinical correctness. Consequently, doctors face pressure to maintain high ratings in order to sustain patient demand, yet they lack clear guidance on which aspects of their communication to change. This creates both an opportunity and a challenge: feedback can incentivize better patient communication and help platforms improve quality, but it can also be difficult for doctors to interpret and act upon, especially when it is negative.

This work attempts to address a key question in telemedicine platforms: \textit{which parts of a doctor's written response are most associated with higher patient satisfaction, and do these patterns differ across patient and clinician subgroups?}

A further motivation for this study is the linguistic context. We analyse interactions written in \textit{Romanian}~\cite{niculae2025retrieval,rogoz2025medqaro}, a low-resource language~\cite{nigatu-etal-2024-zenos}. 
Most \emph{Natural Language Processing} (NLP) research in healthcare has focused on English and other high-resource languages, with relatively limited work on Romanian. As a result, key NLP resources for Romanian, such as clinical-domain models and standardised preprocessing pipelines, remain less developed and less thoroughly validated~\cite{DTVBMzock,dumitrescu2018rowordnet}. To keep the analysis interpretable and realistic, we therefore rely primarily on language-agnostic, surface-level measurements (e.g., length and structure) and incorporate Romanian-specific resources through the Romanian \emph{Linguistic Inquiry and Word Count} (LIWC) lexicon~\cite{duduau2022development,pennebaker2001linguistic}.

% The data undergoes irreversible de-identification to ensure that individual patients cannot be re-identified under any circumstances. Instead of retaining personally identifiable information, patients are categorized into broad, non-identifiable groups based on general characteristics and prior activity. These groupings include distinctions such as new versus returning users, whether they have previously provided positive feedback or no feedback, and gender. By relying solely on aggregated and non-specific attributes, this approach preserves data utility while maintaining strict privacy protection.

In this work, we use completely anonymised data consisting of Romanian-language text-based telemedicine consultations, to study which factors correlate with patient satisfaction, as reflected in post-consultation ratings. 
The data underwent irreversible de-identification to ensure that individuals could not be re-identified. Instead of retaining personally identifiable information, individuals were categorized into broad, non-identifiable groups based on general characteristics and prior activity. These groupings include distinctions such as new versus returning users, whether they have previously provided positive feedback or no feedback, and gender. By relying solely on aggregated and non-specific attributes, this approach preserves data utility while maintaining strict privacy protection.
We treat feedback as the result of multiple interacting components: 

\begin{enumerate}[label=(\roman*)]
    \item Patient group information
    \item Clinician public profile and platform activity
    \item Question metadata (e.g., daytime interval, day of week, chosen medical specialty)
    \item Linguistic properties of the doctor response
\end{enumerate}

Importantly, we do not attempt to assess medical correctness, we strictly focus on observable, surface-level attributes of doctors' responses, features that can be measured and acted upon. We also do not include the patient question text in our analysis.

%poate aici sa zicem mai exact ca se refera la doctor response.
We extract features related to readability and form (e.g., length and structure), psycholinguistic signals (e.g., LIWC categories), and metadata linked to the patient groups, doctors and questions. We then train a gradient-boosting model to predict patient feedback, choosing it as a strong and simple baseline that supports post-hoc interpretability. To understand what the model learns, we use SHAP \cite{shap} to quantify global feature importance and provide local explanations for individual interactions, highlighting which response characteristics most strongly influence predicted satisfaction and how these effects vary across patient and doctor subgroups.

%% file: sections/2.related.tex
Large language models (LLMs) are increasingly adapted to medical settings via pre-training, fine-tuning, and prompting on clinical corpora \cite{zhou2023survey}. Much of the literature evaluates exam-style or multiple-choice QA, where strong performance can be achieved under appropriate adaptation strategies \cite{zhou2023survey}. More recent "copilot" frameworks aim to support broader clinical workflows without task-specific fine-tuning \cite{ren2024healthcare}, but open-ended, real-world interaction settings remain less studied.

Patient-facing deployment depends not only on correctness but also on perceived reliability and empathy. AI-labeled medical advice can be rated as less trustworthy or empathetic, lowering willingness to follow it even when content is held constant \cite{reis2024influence,hohenstein2023artificial}. Conversely, users judge AI responses as comparable to clinicians despite low accuracy \cite{shekar2024people}. Reliability can also be user-dependent: model behaviour may degrade in public-facing use, plausibly because outputs adapt to the quality and specificity of user inputs \cite{anthropic-economic-index,bean2025reliability}.

Beyond LLMs, medical humanities and sociolinguistics emphasize that patients value empathy, clarity, attention to lived experience and appropriate boundary-setting \cite{stergiopoulos2023makes}. A well presented medical advice is more likely to have a positive impact \cite{martin2005challenge} and to attract positive reviews, directly impacting doctor reputation \cite{street2009does,sargeant2008understanding}. Quantitative work links physicians' language to outcomes in online consultations, for example positive-emotion and instrumental language correlating with higher returns \cite{geng2024association}. Classic analyses also report systematic gender differences in doctor--patient dialogue \cite{meeuwesen1991verbal}, motivating subgroup analyses when studying satisfaction. 

Rather than asking whether LLMs provide correct answers, we study what patients reward in \emph{human} clinician writing. We measure which observable properties of doctor replies (surface form, readability proxies, psycholinguistic features) are associated with positive feedback. This complements Romanian-focused efforts on optimizing LLM-assisted doctor communication \cite{niculae-etal-2025-dr}. Our goal is not to assess clinical correctness, but to provide an interpretable view of communication cues that correlate with satisfaction.

%% file: sections/3.method.tex
We study anonymised text-based consultations. After an interaction, which consists of a patient question and a doctor response, patients may leave structured feedback (thumbs-up / thumbs-down). In practice most interactions receive no explicit rating and explicit negative feedback is rare. We therefore model patient satisfaction as a binary prediction task, treating thumbs-up as the positive class and grouping thumbs-down and missing feedback into the other class. As such, the task is to predict whether a doctor response will receive positive feedback or not. This formulation roughly corresponds to the \emph{Positive-Unlabeled learning} (PU-learning) \cite{bekker2020learning} setup. The patient and doctor usually do not have a prior connection and each interaction is stand-alone. Patients can either choose which doctor to send their question to, or they can choose to have the doctor assigned by the platform.
%The dataset is approximately balanced. 
Although completely anonymised, the data is privately owned and cannot be released.

A particular aspect of the studied text consultations is that doctors may address clarification questions before providing an answer. We retain this structure and distinguish between (i) \emph{Simple Questions} and (ii) \emph{Questions with Clarifications} (a clarification step occurs), which we also use for stratified reporting.

\noindent \textbf{Feature extraction}
To keep the analysis interpretable and robust in a lower-resource language setting, we focus primarily on surface-level, mostly language-agnostic features, and add Romanian-specific resources where available. We extract features from four sources: \textit{(i)} patient group information -- new vs returning user, prior activity (e.g., previous thumbs-up), \textit{(ii)} public doctor profile -- metadata (e.g., workplace, education), platform activity and historical feedback rate, 
\textit{(iii)} question metadata (e.g., daytime interval, day of week, chosen medical specialty), 
%-- basic structure and readability proxies (e.g., length), plus LIWC psycholinguistic categories, and 
\textit{(iv)} doctor response text -- binned quantities analysed in ranges for length / structure (number of characters, tokens, punctuation such as question marks), concreteness proxies (e.g., number of digits), and lexicon-based markers (e.g., politeness, hedging, recommendation cues), as well as LIWC features. We compute a fixed feature vector, which is used for predictive modelling.

%To keep the analysis interpretable and robust in a lower-resource language setting, we focus primarily on surface-level, mostly language-agnostic features, and add Romanian-specific resources where available. We extract features from four sources: {\color{red}\textit{(i)} patient profile -- platform tenure, activity, and prior behaviour (e.g., historical thumbs-up rate, number of posed questions per year), }\textit{(ii)} public doctor profile -- metadata (e.g., workplace, education) and platform activity and historical feedback rate. \textit{(iii)} question metadata and patient text -- basic structure and readability proxies (e.g., length), plus LIWC psycholinguistic categories, and \textit{(iv)} doctor response text -- length / structure (characters, tokens, punctuation counts such as question marks), concreteness proxies (e.g., number of digits), and lexicon-based markers (e.g., politeness, hedging, recommendation cues), as well as LIWC features. We compute a fixed feature vector, which is used for predictive modelling.

\noindent \textbf{Prediction model and evaluation protocol}
We train a gradient-boosted decision tree classifier (CatBoost) for binary feedback prediction. To reduce temporal leakage, we use a time-based split: training uses earlier interactions and evaluation is performed on later interactions, for a 80:20 train:test split. We report performance both on the full validation set and on relevant subsets.

\noindent \textbf{Model interpretation and subgroup analyses}
To understand which factors drive predictions, we compute SHAP values on the trained model. We summarize global importance with mean absolute SHAP values and use local explanations to inspect individual decisions. In addition, to probe whether associations differ across populations, we perform subgroup correlation analyses between selected text features and feedback, stratifying by patient gender (\emph{patient preferences}) and doctor gender (\emph{doctor behaviour}), and report the resulting patterns descriptively.

%% file: sections/4.results.tex
\input{tables/classifier-performance}

\begin{figure}[hbt!]
    \centering
    \includegraphics[width=\linewidth]{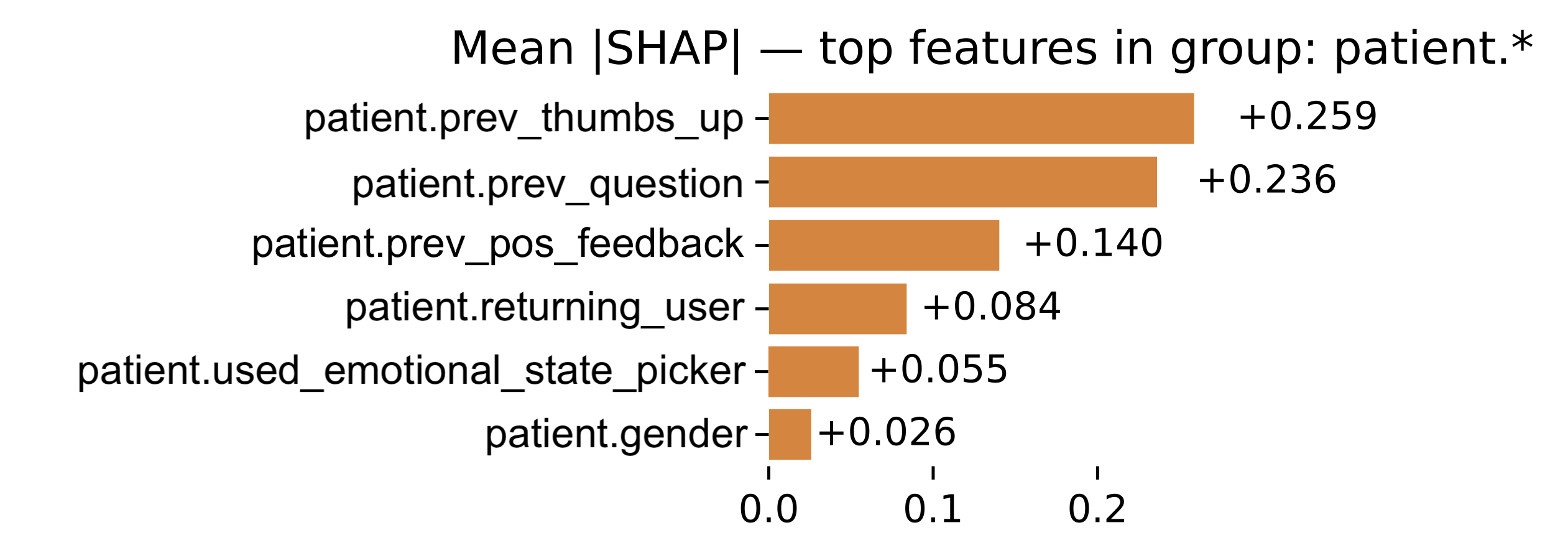}
    \includesvg[width=\linewidth]{images/doctor_importance.svg}
    % \includesvg[width=0.49\linewidth]{images/patient_text_importance.svg}
    \includesvg[width=\linewidth]{images/doctor_text_importance.svg}
    \caption{Mean absolute SHAP values for the CatBoost feedback classifier. Larger values indicate greater impact on predictions (magnitude only; SHAP sign is not shown). Quantities are binned and are not processed in absolute numbers.}
    \label{fig:importances}
\end{figure}

\begin{figure*}[hbt!]
    \centering
    \includesvg[width=0.49\linewidth]{images/patient_preferences_correlation.svg}
    \includesvg[width=0.49\linewidth]{images/doctor_behaviour_correlation.svg}
    \caption{Pearson correlations between selected doctor-response text features and positive feedback, aggregated over all interactions. Positive values indicate features that tend to co-occur with thumbs-up feedback, while negative values indicate features that tend to co-occur with thumbs-down or missing feedback. Quantities are binned and are not processed in absolute numbers.}
    \label{fig:behaviour-text-correlation}
\end{figure*}

% \begin{figure}[hbt!]
%     \centering
%     \includesvg[width=0.75\linewidth]{images/male-female-doctors.svg}
%     \caption{Pearson correlations between doctor-response text features and positive feedback, computed separately by doctor gender (\emph{Doctor Behaviour}) and by patient gender (\emph{Patient Preferences}). Correlations are modest overall, but patterns are broadly consistent across groups and highlight which linguistic markers associate with satisfaction.}
%     \label{fig:male-female-doctor}
% \end{figure}

In Table \ref{tab:subset-eval-metrics}, we show classifier performance across subgroups. Since the classifier is properly calibrated (calibration error of 1.5\%), high confidence predictions have high performance in terms of ROC-AUC score. Results are also better for clarification-type questions: clarification questions may function as a visible signal of attention and engagement.

\noindent \textbf{Correlation analysis by doctor and patient gender.}
Figure~\ref{fig:behaviour-text-correlation} shows Pearson correlations between doctor-text features and binary feedback, computed separately by doctor gender (\emph{Doctor Behaviour}) and by patient gender (\emph{Patient Preferences}). Across all splits, the largest positive correlations are for politeness markers and hedging in the doctor response text, consistent with prior work \cite{stergiopoulos2023makes}. Smaller positive correlations appear for features that often signal concrete, actionable replies, such as the number of digits, explicit recommendations and question marks. The most consistent negative correlation is for lexical diversity, with additional negative correlations for several LIWC categories (e.g., \texttt{liwc\_work}, \texttt{liwc\_affiliation}, \texttt{liwc\_social}). Higher lexical diversity can reflect more complex, less standardized phrasing, which may increase cognitive load; conversely, simpler, more formulaic language may read as clearer and safer \cite{stergiopoulos2023makes}.

The overall pattern is similar across the two gender groups, suggesting that both male and female patients tend to respond positively to polite, cautious and concrete messages. Differences by patient gender are small. Differences by doctor gender are more visible: for several of the top positive features (especially politeness, hedging and number of digits), correlations are larger within the subset of replies written by female doctors than within those written by male doctors. Correlation magnitudes are modest overall (roughly $|r|\approx 0.02$--$0.15$), so these results should be treated as descriptive. 

\noindent \textbf{Model explanations using SHAP.}
Figure~\ref{fig:importances} reports mean absolute SHAP values (feature importance). Overall, the model relies most on \emph{patient group information} features. The largest contributions come from the returning users feature and the previous thumbs-up feature. These features mainly act as strong priors: patients that previously gave positive feedback are more predictable.

Doctor-level features contribute less than patient group information. The top doctor predictors are previously received thumbs-up rate, doctor workplace, number of previously helped patients, and number of years on the platform, along with past positive / negative feedback rates. This suggests the model captures persistent differences between clinicians and settings (e.g., workplace) beyond the doctor response.

Doctor response text features have smaller mean $|\mathrm{SHAP}|$ values than patient groups and doctor public profiles features, indicating that metadata dominates the prediction. The most important doctor-text signals include number of questions, number of digits present, and several LIWC categories (e.g., \texttt{liwc\_work}, \texttt{liwc\_negemo}, \texttt{liwc\_time}), as well as length proxies (\texttt{n\_characters}, \texttt{n\_tokens}). These patterns are consistent with feedback being associated with replies that are more interactive (questions/clarifications), more concrete (numbers) and more detailed (longer responses), confirming prior work \cite{geng2024association}. Digits often encode dosage, timing, thresholds or step-by-step plans, which are surface markers of actionability. Satisfaction signals are identity- and expectation-loaded, as ratings reflect the rater’s baseline tendency and the clinician’s reputation/context more than the response. This is consistent with "perception-first" evaluation dynamics discussed in prior work \cite{shekar2024people}, and motivates controlling for priors before making communication recommendations.

\begin{figure*}[hbt!]
    \centering
    \includegraphics[width=\linewidth]{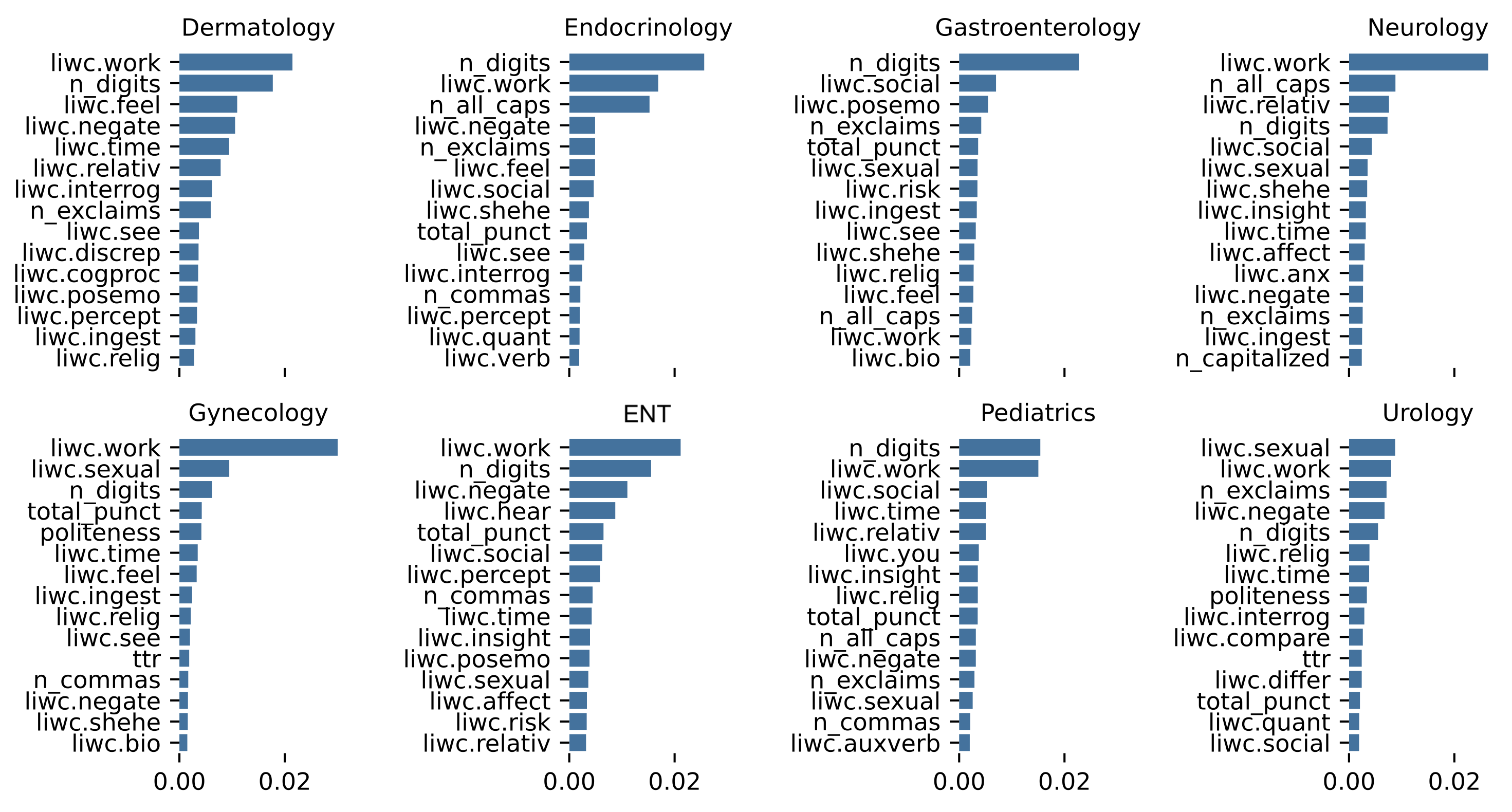}
    \caption{Mean absolute SHAP values for the most important doctor text features, computed separately for each medical question specialty. Bars show the response-text features with the largest contribution magnitude to the feedback prediction model within each specialty. Values indicate feature importance magnitude only and do not show effect direction. Quantities are binned and are not processed in absolute numbers.}
    \label{fig:shap-per-category}
\end{figure*}

Figure \ref{fig:shap-per-category} shows that, when restricting the analysis to doctor-response text and stratifying by medical specialty, the most predictive cues become more specialty-specific. Some features recur across categories, especially digits, time-related language, and task-oriented terms, suggesting that patient feedback favors responses that are concrete, procedural and actionable. At the same time, several categories display clinically intuitive lexical patterns: dermatology emphasizes perceptual and symptom-description language; endocrinology is strongly driven by numeric content; gastroenterology highlights ingestion- and risk-related terms; ENT uniquely surfaces hearing-related language; pediatrics gives more weight to social and second-person terms; and gynecology/urology place greater emphasis on sexual and politeness-related language. Overall, results suggests that positive feedback is associated not only with general clarity, but with responses whose language is appropriately matched to the communicative demands of the medical specialty.

%% file: tables/classifier-performance.tex
\begin{table}[t]
\centering
\resizebox{\linewidth}{!}{
    \begin{tabular}{llrrrrr}
    \textbf{Subset} & \textbf{Split} & \textbf{ROC-AUC} & \textbf{PR-AUC} & \textbf{F1} & \textbf{Precision} & \textbf{Recall} \\
    \hline
    Simple Questions      & Val       & 0.6780 & 0.6865 & 0.6206 & 0.6206 & 0.6209 \\
    Only Clarifications & Val       & 0.7153 & 0.5857 & 0.6367 & 0.6411 & 0.6493 \\
    All Questions       & Val       & 0.7192 & 0.6313 & 0.6518 & 0.6542 & 0.6577 \\
    \midrule
    New Patients & Val & 0.6479 & 0.6003 & 0.5804 & 0.6061 & 0.5963 \\
    Returning Patients & Val & 0.7656 & 0.6533 & 0.6951 & 0.6968 & 0.6937 \\
    \midrule
    Simple Questions      & Val High-conf & 0.6951 & 0.6138 & 0.6456 & 0.7724 & 0.6416 \\
    Only Clarifications & Val High-conf & 0.8348 & 0.6542 & 0.7751 & 0.7960 & 0.7603 \\
    All Questions       & Val High-conf & 0.8357 & 0.6987 & 0.7889 & 0.8001 & 0.7806 \\
    \end{tabular}
}
\caption{Feedback prediction performance across question subsets and patient groups (New vs.\ Returning), reported on the validation split and a high-confidence subset.}
\label{tab:subset-eval-metrics}
\end{table}

%% file: sections/5.conclusions.tex
We analysed anonymised Romanian text-based telemedicine consultations to examine which characteristics of written clinician replies are associated with positive patient feedback.
The main finding is that satisfaction is driven largely by stable priors (patient group information and doctor activity), so predicting feedback mostly captures who tends to rate positively and who tends to receive positive ratings. Still, the response text contains smaller but actionable signals: politeness and hedging are consistently associated with positive feedback, as are concrete, action-oriented cues (e.g., digits and recommendations), while higher lexical diversity correlates negatively. Presumably, more high-level and semantic response features tailored to this specific use-case could be discovered automatically using feature discovery methods \cite{cosma2026automaticpromptoptimizationdatasetlevel}. 
The results suggest that patient feedback should be interpreted primarily as a signal of clear, empathetic written communication, rather than as a proxy for medical accuracy. 

Future work should focus on modelling higher-level, semantic features of the doctor responses, such as response clarity and whether the response contains adequate explanations where appropriate. These aspects are not easily captured through lexicon-based approaches and may require the use of pretrained language models \cite{cosma2026automaticpromptoptimizationdatasetlevel}. 